\documentclass{article}
\usepackage{nips01,times}
\usepackage{graphicx}
\usepackage{subfigure}
\usepackage{psfig}

\title{The Dynamics of AdaBoost Weights \\ Tells You What's Hard to Classify}

\author{B. Caprile\\
ITC-irst \\
I-38050 Povo, Trento\\
Italy\\
{\it caprile@itc.it} \\
\And
C. Furlanello \\
ITC-irst \\
I-38050 Povo, Trento\\
Italy\\
{\it furlan@itc.it} \\\\
\And
S. Merler\\
ITC-irst \\
I-38050 Povo, Trento\\
Italy\\
{\it merler@itc.it} \\\\
}

\begin{document}
\maketitle

\bibliographystyle{plain}

\newcommand{\REM}[1]{
{\bf #1}
}

\newcommand{\Ada}{AdaBoost
}

\newcommand{\Adaa}{{\tt AdaBoost} algorithm
}

\begin{abstract}
The dynamical evolution of weights in the \Ada algorithm contains
useful information about the r{\^o}le that the associated data points
play in the built of the \Ada model. In particular, the dynamics
induces a bipartition of the data set into two (easy/hard)
classes. Easy points are ininfluential in the making of the model,
while the varying relevance of hard points can be gauged in terms of
an entropy value associated to their evolution. Smooth approximations
of entropy highlight regions where classification is most
uncertain. Promising results are obtained when methods proposed are
applied in the Optimal Sampling framework.
\end{abstract}

\begin{section}{Introduction}

In this paper we investigate the boosting weight dynamics induced by
classification procedures of the AdaBoost family
\cite{FreSch97,SchFreBarLee98}, and show how it can be exploited to
for highlighting points and regions of uncertain
classification. Friedman et al. \cite{FriHasTib00} proposed to analyze
and trim the distribution of weights over a training sample in order
to reduce computation without sacrificing accuracy. Here, we focus
instead on tracking the dynamics of the boosting weight of individual
points. By introducing the notion of entropy of the weight evolution,
we can clarify the notions of ``easy'' and the ``hard'' points as the
two types of weight dynamics being observed: in particular, in
different classification tasks and with different base models it is
found that a group of points may be selected which have very low
(ideally, zero) entropy of weight evolution: the easy points. In this
framework, we can answer questions as: do easy point play any role in
building the AdaBoost model? For hard points, can different degrees
of ``hardness'' be identified which account for different degrees of
classification uncertainty? Do easy/hard points show any preference about
where to concentrate? The first two questions are clearly connected to
equivalent results in the framework of Support Vector Machines: in a
number of experiments, hard points are
found indeed mostly nearby the classification boundary.  In the second
part of this paper, the smooth approximation (by kernel regression) of
the weight entropy at training data is proposed as an indicator
function of classification uncertainty, thereby obtaining a region
highlighting methodology. As a natural application, 
a strategy for optimal sampling in classification tasks was implemented:
compared with uniform random sampling, the entropy-based strategy is
clearly more effective. Moreover, it compares favorably with an
alternative margin-based sampling strategy. 

\end{section}

\begin{section}{The Dynamics of Weights}
\label{sec:dynamics}

In the present section, the dynamics that the \Ada algorithm sets over
the weights is singled out for study. In particular, the intuition is
substantiated that the evolution of weights yields information about
the varying relevance that different data points have in the built of
the \Ada model. 

Let $D \equiv \{{\bf x}_{i}, y_{i}\}_{i=1}^{N}$ be a two-class set of
data points, where the ${\bf x}_{i}$s belong to a suitable region,
$X$, of some (metric) feature space, and $y_{i}$ takes values in $\{1,
-1\}$, for $1 \leq i \leq N$. The \Ada algorithm iteratively builds a
class membership estimator over $X$ as a thresholded linear
superposition of different realizations, $M_{k}$, of a same base
model, $M$. Any model instance, $M_{k}$, resulting from training at
step $k$ depends on the values taken at the same step by a set of $N$
numbers (in the following, the {\em weights}), ${\bf w} = w_{1}, \dots
w_{N}$ -- one for each data point. After training, weights are
updated: those associated to points misclassified by the current model
instance are increased, while decreased are those for which the
associated point is classified correctly. An interesting variant of
this basic scheme consists in training the different realizations of
the base model, not on the whole data set, but on Bootstrap replicates
of it \cite{Qui96}. In this second scheme, samplings are extracted
according to the discrete probability distribution defined by the
weights associated to data points, normalized to sum one.

In Fig. \ref{fig:weights-traces-and-histograms}a the plots are
reported of the evolution of the weights associated to 3 data points
when the \Ada algorithm is applied to a simple binary classification
task on synthetic two-dimensional data (experiment A-{\tt Gaussians}
as described in Sec. \ref{subsec:appendix-data-a}). Except for
occasional bursts, the weight associated to the first point goes
rapidly to zero, while the weights associated to the second and third
point keep on going up and down in a seemingly chaotic fashion. Our
experience is that these two types of behaviour are not specific of
the case under consideration, but can be observed in any \Ada
experiment. Moreover, {\em tertium non datur}, i.e., no other
qualitative behaviour is observed (as, for example, that some weight
tends to a strictly positive value).

\begin{subsection}{Easy Vs. Hard Data Points}
\label{easy-hard-data-points}	

\begin{figure*}[ht]
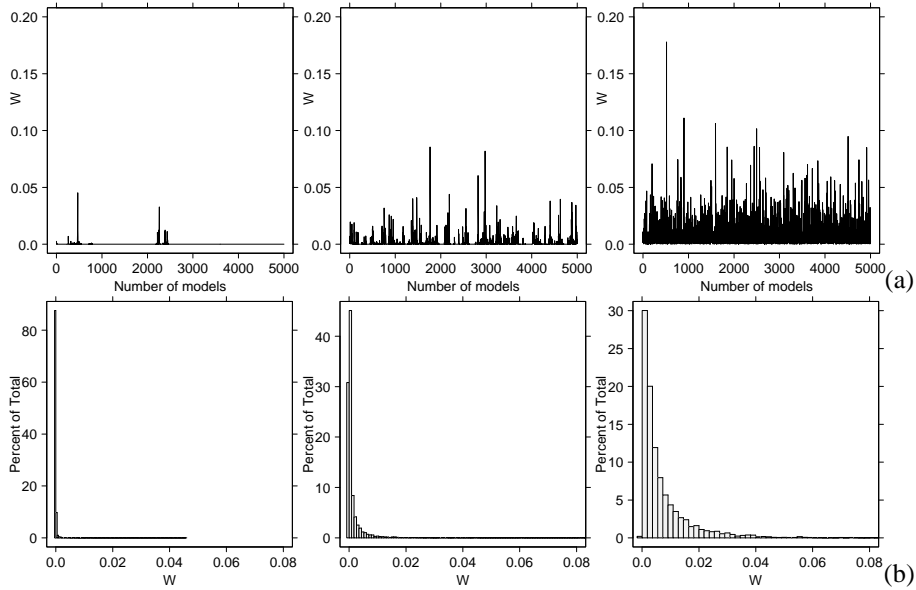

  \begin{center} 
    \leavevmode
    \psfig{figure=gaussian-5000-weights-trace-1.epsi,width=0.3\textwidth}
    \psfig{figure=gaussian-5000-weights-trace-2.epsi,width=0.3\textwidth}
    \psfig{figure=gaussian-5000-weights-trace-3.epsi,width=0.3\textwidth}(a)
    \psfig{figure=gaussian-5000-histogram-1.epsi,width=0.3\textwidth}
    \psfig{figure=gaussian-5000-histogram-2.epsi,width=0.3\textwidth}
    \psfig{figure=gaussian-5000-histogram-3.epsi,width=0.3\textwidth}(b)
    \caption{{\em Evolution of weights in the \Ada algorithm. (a)
    The evolutions over 5000 steps of the \Ada algorithm are reported
    for the weights associated to 3 data points of experiment {\rm
    A-{\tt Gaussians}}. From left to right: an ``easy'' data point
    (the weight tends to zero), and two ``hard'' data points (the
    weight follows a seemingly random pattern). (b) The corresponding
    frequency histograms.}}
    
    \label{fig:weights-traces-and-histograms}
  \end{center}
\end{figure*}

The hypothesis therefore emerges that the \Ada algorithm set a
partition of data points into two classes: on one side the points
whose weight tends rapidly to zero; on the other, the points whose
weight show an apparently chaotic behaviour. In fact, the hypothesis is
perfectly consistent with the rationale underlying the \Ada algorithm:
weights associated to those data points that several model instances
classify correctly even when they are {\em not} contained in the
training sample follow the first kind of behaviour. In practice
independently of which bootstrap sample is extracted, these points are
classified correctly, and their weight is consequently decreased and
decreased. We call them the ``easy'' points. The second type of
behaviour is followed by the points that, when not contained in the
training set, happen to be often misclassified. A series of
misclassifications makes the weight associated with any such point
increase, thereby increasing the probability for the point to be
contained in the following bootstrap sample. As the probability
increases and the point is finally extracted (and classified
correctly), its weight is decreased; this in turn makes the point less
likely to be extracted -- and so forth. We call this kind of points
``hard''.

In Fig. \ref{fig:weights-traces-and-histograms}b, histograms are
reported of the values that the weights associated to the same 3 data
points of Fig. \ref{fig:weights-traces-and-histograms}a take over the
same 5000 iterations of the \Ada algorithm. As expected, the histogram
of (easy) point 1 is very much squeezed towards zero (more than 80\%
of weights lies below $10^{-6}$). Histograms of (hard) points 2 and 3
exhibit the same Gamma-like shape, but differ remarkably for what
concerns average and dispersion. Naturally, the first question is
whether any limit exists for these distributions. For each data point,
two unbinned cumulative distributions were therefore built by taking
the weights generated by the first 3000 steps of the \Ada algorithm,
and those generated over the whole 5000 steps. The same-distribution
hypothesis was then tested by means of the Kolmogorov-Smirnov (KS)
test \cite{PreTeuVetFla92}. Results are reported in
Fig. \ref{fig:mean-vs-entropy-ks-test-and-histogram}a, where
$p$-values are plotted against the mean value of all 5000 values. It
is interesting to notice that for mean values close to 0 (easy points)
the same-distribution hypothesis is always rejected, while it is
typically not-rejected for higher values (hard points). It seems that
easy points may be confidently identified by simply considering the
average of their weight distribution. A binary LDA classifier was
therefore trained on the data of
Fig. \ref{fig:mean-vs-entropy-ks-test-and-histogram}a. By setting a
$p$-value threshold equal to 0.05, the resulting {\em precision} (the
complement to 1 of the fraction of false negative) was equal to 0.79
and {\em recall} (the complement to 1 of the fraction of false
positive) was equal to 0.96.

\end{subsection}

\begin{subsection}{Entropy}
\label{subsec:entropy}

Can we do any better at separating easy points from hard ones? For
hard points, can different degrees of ``hardness'' be identified which
account for different degrees of classification uncertainty? What we
are going to show is that by associating a notion of {\em entropy} to
the evolutions of weights both questions can be answered in the
positive. To this end, the interval $[0,1]$ is partitioned into $L$
subintervals of length $1/L$, and the entropy value is computed as
$\sum_{i=1}^{L} f_{i}~log_{2}~ f_{i}$, where $f_{i}$ is the relative
frequency of weight values falling in the $i$-th subinterval ($0~
log_{2}~ 0$ is set to $0$). For our cases, $L$ was set to 1000.

\begin{figure*}[ht]
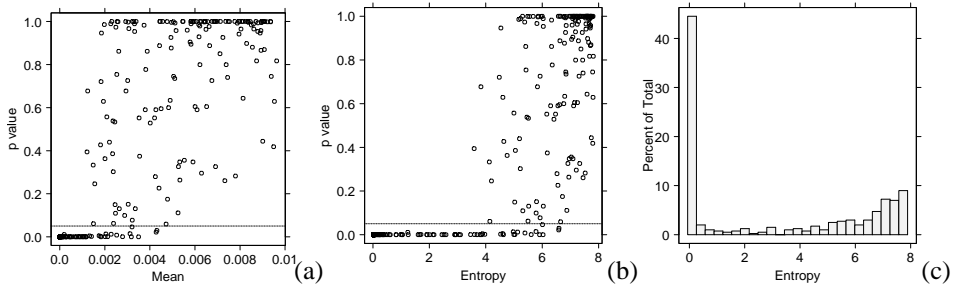

  \begin{center}
    \leavevmode
	\psfig{figure=ks-test-mean.epsi,width=0.29\textwidth}(a)
    \psfig{figure=ks-test-entropy.epsi,width=0.29\textwidth}(b)
    \psfig{figure=entropy-histogram.epsi,width=0.29\textwidth}(c)
    \caption{{\em Separating easy form hard points. (a) $p$-values of
    the KS test Vs. mean values of frequency histograms. (b)
    $p$-values of the KS test Vs. entropy of frequency histograms. As
    in (a), the horizontal line marks the threshold value for the LDA
    classifier. (c) Histogram of entropy values for the 400 data
    points of experiment {\rm A-{\tt Gaussians}}.}}
    
    \label{fig:mean-vs-entropy-ks-test-and-histogram}
  \end{center}
\end{figure*}

Qualitatively, the relationship between entropy and $p$-values of the
KS test is similar to the one holding for the mean
(Fig. \ref{fig:mean-vs-entropy-ks-test-and-histogram}a-b). Quantitatively,
however, a difference is observed, since the LDA classifier trained on
these data performs much better in precision and slightly worse in
recall (respectively, 0.99 and 0.90, as compared to 0.79 and
0.96). This implies that the class of easy points can be identified
with higher confidence by using the entropy in place of the mean value
of the distribution. Further support to the hypothesis of a bipartite
(easy/hard) nature of data points is gained by observing the frequency
histogram of entropies for the 400 points of experiment A-{\tt
Gaussians} (Fig. \ref{fig:mean-vs-entropy-ks-test-and-histogram}c),
from which two groups of data points emerge as clearly separated. The
first is the zero entropy group of easy points, and the second is the
group of hard points.

Do easy/hard points show any preference about where to concentrate?
In Fig. \ref{fig:using-entropy}a hard and easy points are shown as
determined for the experiment A-{\tt Sin} (see
Sec. \ref{subsec:appendix-data-a} for details). Hard points are mostly
found nearby the two-class boundary; yet, their density is much lower
along the straight segment of the boundary (where the boundary is
smoother), and appear therefore to concentrate where the
classification uncertainty is highest. Easy points to the
opposite. Considering that easy points stay well clear of the boundary
(i.e., hard points typically interpose between them and the boundary),
what one may then question is whether they play any r{\^o}le in the
built of the \Ada model. The answer is no. In fact, the models built
disregarding the easy points are practically the same as the models
built on the complete data set. In the experiment of
Fig. \ref{fig:using-entropy} only the $0.55\%$ of $10000$ test points
were classified differently by the two models, as contrasted to
reduction of the training set from $400$ to only $111$ (hard)
points. 

\end{subsection}

\begin{subsection}{Smoothing the Entropy}
\label{subsec:extending-entropy}

In the previous section, the entropy of the weight frequency histogram
was introduced as an indicator of the uncertainty of classifying the
associated data point as belonging to class $-1$ or $1$. By defining a
smooth approximation to the punctual entropy values associated to data
points, we now extend the notion of classification uncertainty to the
whole domain of our binary classifier. For simplicity sake, kernel
regression was employed -- i.e., the entropy values at data points are
convolved with a Gaussian kernel of fixed bandwidth \cite{Har90}. In
so doing, a scalar entropy function, $H = H({\bf x})$, is defined on
$A$. In Fig. \ref{fig:using-entropy}b, the grey levels encode the
values of $H$ (increasing from black to white) for the experiment {\rm
A-{\tt Sin}}.

\begin{figure*}[ht]
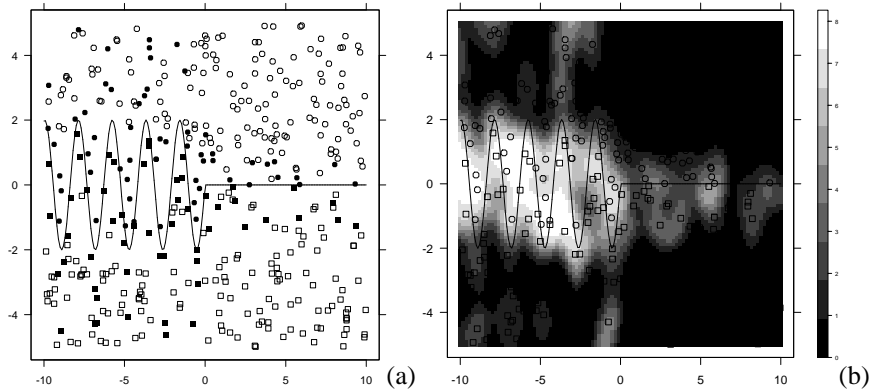

  \begin{center} \leavevmode
    \psfig{figure=sinusoidal-5000-leaving-out-easy-points.epsi,height=0.4\textwidth}(a)
    \psfig{figure=sinusoidal-convolution-0.5.epsi,height=0.4\textwidth}(b)
    \caption{{\em (a) Easy (white) and hard (black) data points of
    experiment A-{\tt Sin} obtained by thresholding the histogram of
    entropy. Squares and circlets express the class. (b) Level-plot of
    the $H$ function. Grey levels encode $H$ values (see scale on the
    right).}}  
\label{fig:using-entropy} 
\end{center}
\end{figure*}

The method appears capable of highlighting regions where
classification turns out uncertain -- due to the distribution of data
points, the morphology of the class boundary or both. Of course,
function $H$ depends on the geometric properties specific of the base
model adopted, and its degree of smoothness depends on the size of the
convolution kernel. It should be noticed, however, that the
bias/variance balance can be controlled by suitably tuning the
convolution parameters. Finally, more sophisticated local smoothing
techniques may be employed as well (e.g., Radial Basis Functions)
which may adapt to directionality, known morphology of the boundary or
local density of sample points.

\end{subsection}

\end{section}

\begin{section}{An Application to Optimal Sampling}
\label{sec:optimal-design}

To illustrate the applicability of notions developed above to
practical cases, we refer to the framework of optimal sampling
\cite{Fed72}. In general, an optimal sampling problem is one in which
a {\em cost} is associated to the acquisition of data points, in such
a way that solving the problem consists not only in minimizing the
classification (or regression) error but also in keeping the sampling
cost as low as possible. A typical setting for this class of problems
is the one in which we start from an assigned set of (sparse) data
points, and we then incrementally add points to the training set on
the basis of certain information extracted from intermediate
results. 


\begin{figure*}[ht]
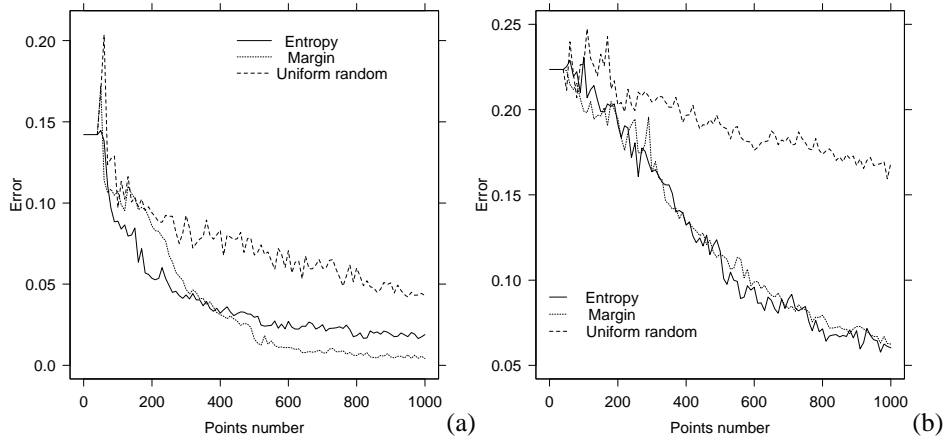

  \begin{center}
    \leavevmode
    \psfig{figure=sinusoidal-incremental-40-1000-x10-error.epsi,width=0.45\textwidth}(a)
    \psfig{figure=spiral-incremental-40-1000-x10-error.epsi,width=0.45\textwidth}(b)
    \caption{{\em Misclassification error as a function of the number
    of training points for the entropy based scheme is compared to
    the uniform random sampling and the margin sampling
    strategy. (a) Experiment {\rm B-{\tt Sin}}. (b) Experiment {\rm B-{\tt Spiral}}.}}
    \label{fig:optimal-sampling-errors} 
  \end{center} 
\end{figure*}
\end{section}

For the experiments reported below, which are based on the same
settings as {\tt Sin} and {\tt Spiral} of
Sec. \ref{subsec:appendix-data-a} (see also
Sec. \ref{subsec:experiment-b} for details), we started from a small
set of sparse two-dimensional binary classification
data. High-uncertainty areas are identified by means of the method
described in Sec. \ref{subsec:extending-entropy}, and additional
training points are chosen in these areas. Assuming a unitary cost for
each new point, performance of the procedure is finally evaluated by
analyzing the sampling cost against the classification error.

In Fig. \ref{fig:optimal-sampling-errors}, two plots are reported of
the classification error as function of the number of training
points. Comparison is made with a blind (randomly uniform) sampling
strategy, and with a specialization of {\em uncertainty sampling
strategy} as recently proposed in \cite{LewCat94}. The latter consists
in adding training points where the classifier is less certain of
class membership. In particular, the classifier was the \Ada model and
the uncertainty indicator was the margin of the prediction.

Results reported in Fig. \ref{fig:optimal-sampling-errors} show that
in both experiments the entropy sampling method holds a definite
advantage on the random sampling strategy. In the first experiment, an
initial advantage of entropy over the margin based sampling is also
observed, but the margin strategy takes over as the number of
samplings goes beyond 400. It should be noticed, however, that the
margin sampling automatically adapts its spatial scale to the
increased density of sampling points, while our entropy method does
not (the size of the convolution kernel is fixed). In fact, in the
experiment {\rm B-{\tt Spiral}}
(Fig. \ref{fig:optimal-sampling-errors}b) where the boundary has a
more complex structure, (and the size of convolution kernel smaller),
1000 samplings are not sufficient for the margin based method to
exhibit an advantage on the entropy method (but the latter looses the
initial advantage exhibited in the first experiment).

\begin{section}{Final Comments}
\label{sec:conclusions}

Within the many possible interpretations of learning by boosting, it
is promising to create diagnostic indicator functions alternative to
margins \cite{SchFreBarLee98} by tracing the dynamics of boosting
weights for individual points. We have used entropy (in the punctual
and then smoothed versions) as a descriptor of classification
uncertainty, identifying easy and hard points, and designing a
specific optimal sampling strategy. The strategy needs to be further
automated, e.g. considering adaptive selection of smoothing parameters
as a function of spatial variability. A direct numerical relationship
with the weights of Support Vector expansions is also clearly needed.
On the other hand, it would be also interesting to associate the
main types of weight dynamics (or point hardness) to the
regularity of the boundary surface and of the noise structure.

\end{section}

\appendix

\begin{section}{Data}
\label{sec:appendix-data}

Details are given on the data employed in experiments of
Sec. \ref{sec:dynamics} and \ref{sec:optimal-design}. Full details and
data are accessible at {\tt http://www.mpa.itc.it/nips-2001/data/}.

\begin{subsection}{Experiment A}
\label{subsec:appendix-data-a}


\begin{description}

        \item[{\tt Gaussians}:] 4 sets of points (100 points each) were
generated by sampling 4 two-dimensional Gaussian distributions,
respectively centered in $(-1.0,0.5)$, $(0.0,-0.5)$, $(0.0,0.5)$ and
$(1.0,-0.5)$. Covariance matrices were diagonal for all the 4
distributions; variance was constant and equal to 0.4. Points coming
from the sampling of the first two Gaussians were labelled with class
$-1$; the others with class $1$.


        \item[{\tt Sin}:] The box in $R^{2}$, $R \equiv
[-10,10]\times[-5,5]$, was partitioned into two class regions $R_{1}$
(upper) and $R_{-1}$ (lower) by means of the curve, $C$ of parametric
equations:

$$ 
C \equiv \left\{
    \begin{array}{rcl}
      x(t) & = & t \\
      y(t) & = & 2 sin(3 t) \mbox{ if } -10 \leq t \leq 0 ; 0 \mbox{
    if } 0 \leq t \leq 10 .\\
    \end{array}
  \right. 
$$

\noindent
400 two-dimensional data were generated by randomly sampling region
$R$, and labelled with either $-1$ or $1$ according to whether they
belonged to $R_{-1}$ or $R_{1}$.

        \item[{\tt Spiral}:] As in the previous case, the idea was to
have a bipartition of a rectangular subset, $S$, of $R^{2}$ presenting
fairly complex boundaries ($S \equiv [-5,5]\times[-5,5]$). Taking
inspiration from \cite{RavInt99}, a spiral shaped boundary was
defined. 400 two-dimensional data were then generated by randomly
sampling region $S$, and were labelled with either $-1$ or $1$
according to whether they belonged to one or the other of the two
class regions.

\end{description}

\end{subsection}

\begin{subsection}{Experiment B}
\label{subsec:experiment-b}

This group of data was generated in support to the optimal sampling
experiments described in Sec. \ref{sec:optimal-design}. More
specifically, two initial data sets, each containing 40 points, were
generated for both the {\tt Sin} and {\tt Spiral} settings by
employing the same procedures as above. At each round of the optimal
sampling procedure, 10 new data points were generated by uniformly
sampling a suitable, high entropy subregion of the domain. Data
points were then labelled according to their belonging to one or the
other of the two class regions.

\end{subsection}

\end{section}

\end{document}